\documentclass[runningheads,orivec]{llncs}
\usepackage[T1]{fontenc}
\usepackage{graphicx}
\usepackage{booktabs}
\usepackage{array}
\usepackage{multirow}
\usepackage{xcolor}
\usepackage{amsmath}
\usepackage{amssymb}
\usepackage{url}
\newcolumntype{L}[1]{>{\raggedright\arraybackslash}p{#1}}
\begin{document}
\title{Order Matters: A Chinese Multi-Panel Meme Benchmark for Vision-Language Reasoning}
\titlerunning{CMPM: Order-Sensitive Chinese Multi-Panel Memes}
\author{Haihan Li\inst{1} \and Haihao Li\inst{2} \and
Zhenfei Xu\inst{1} \and Jize Qian\inst{1} \and
Yubo Xie\inst{3}\thanks{Corresponding author.}}
\authorrunning{H. Li et al.}
\institute{Shanghai Maritime University, Shanghai, China
\and
Dalian Maritime University, Dalian, China
\and
Oxford Suzhou Centre for Advanced Research}
\maketitle
\begin{abstract}
Many multimodal tasks depend on how visual elements are ordered and composed, not only on recognizing them in isolation.
Internet memes are a compact case of this problem: their punchline often depends on a constrained reading order and cross-panel visual--textual cues.
While large vision-language models (LVLMs) show strong performance on single-image understanding, it remains unclear whether they can perform \emph{sequence-aware} reasoning over structured meme layouts, especially in Chinese social media.
We introduce \textbf{CMPM}, a Chinese Multi-Panel Meme benchmark with 1{,}214 annotated samples covering five structural types, ordering dependency, panel-order constraints, and optional comment context.
We formulate a two-layer evaluation: Task~1 probes structure typing and order-sensitive panel sequencing (with a context ablation setting), and Task~2 evaluates Chinese meme explanation generation with \emph{human ratings} on five 1--3 Likert dimensions (visual, panel, humor, context, and faithfulness).
We benchmark five representative LVLMs under a unified protocol.
Results indicate that canonical-display accuracy is not by itself evidence of order understanding: the primary shuffled condition produces a sharp accuracy drop, revealing a persistent gap in order-sensitive multimodal reasoning.
Task~2 preferences place Gemini~3.1~Pro and GPT-5.5 above the open models, while comment context yields only a small and mixed Core4 gain.
Code and data will be released upon acceptance.

\keywords{Multi-panel memes \and Multimodal reasoning \and Human evaluation}
\end{abstract}
\section{Introduction}
\label{sec:intro}

Multimodal understanding often depends on how visual elements are arranged, sequenced, and jointly interpreted, not just on recognizing objects or text in isolation.
Memes are a pervasive form of multimodal communication that combine images, overlaid text, and cultural references to convey humor, irony, or social commentary~\cite{sharma2020semival,kiela2020hateful}.
Beyond the classic single-image template, Chinese-language social platforms commonly circulate \emph{multi-panel} memes---comic-strip layouts, chat screenshots, and comparison grids---in which meaning emerges from the \emph{sequence} of panels and their inter-panel relations.
Reordering panels can destroy the punchline, invert a contrast, or break a conversational turn, making reading order a first-class variable for multimodal understanding.

Large vision-language models (LVLMs) such as LLaVA~\cite{liu2023llava}, Qwen-VL~\cite{bai2023qwen}, InternVL~\cite{chen2024internvl}, and proprietary systems have advanced image captioning, visual question answering, and document understanding.
Existing meme benchmarks primarily target classification, hate/offense detection, or single-image explanation~\cite{kiela2020hateful,suryawanshi2020multimodal,he2025memereacon}.
Many of these tasks can be solved through local cue matching: recognizing salient people, objects, or text and associating them with a label or likely interpretation without reconstructing global panel relations. They therefore need not test setup-to-punchline dependencies or distinguish a valid reading order from a shuffle.
Text-only Chinese meme resources such as CHIME~\cite{xie2025chime} evaluate cultural phrase understanding, but do not address panel structure or visual order.
To our knowledge, existing resources do not provide a dedicated testbed for \emph{order-sensitive} multimodal reasoning on Chinese multi-panel memes.

\begin{figure}[!h]
\centering
\includegraphics[width=\textwidth,trim=0pt 0pt 0pt 55pt,clip]{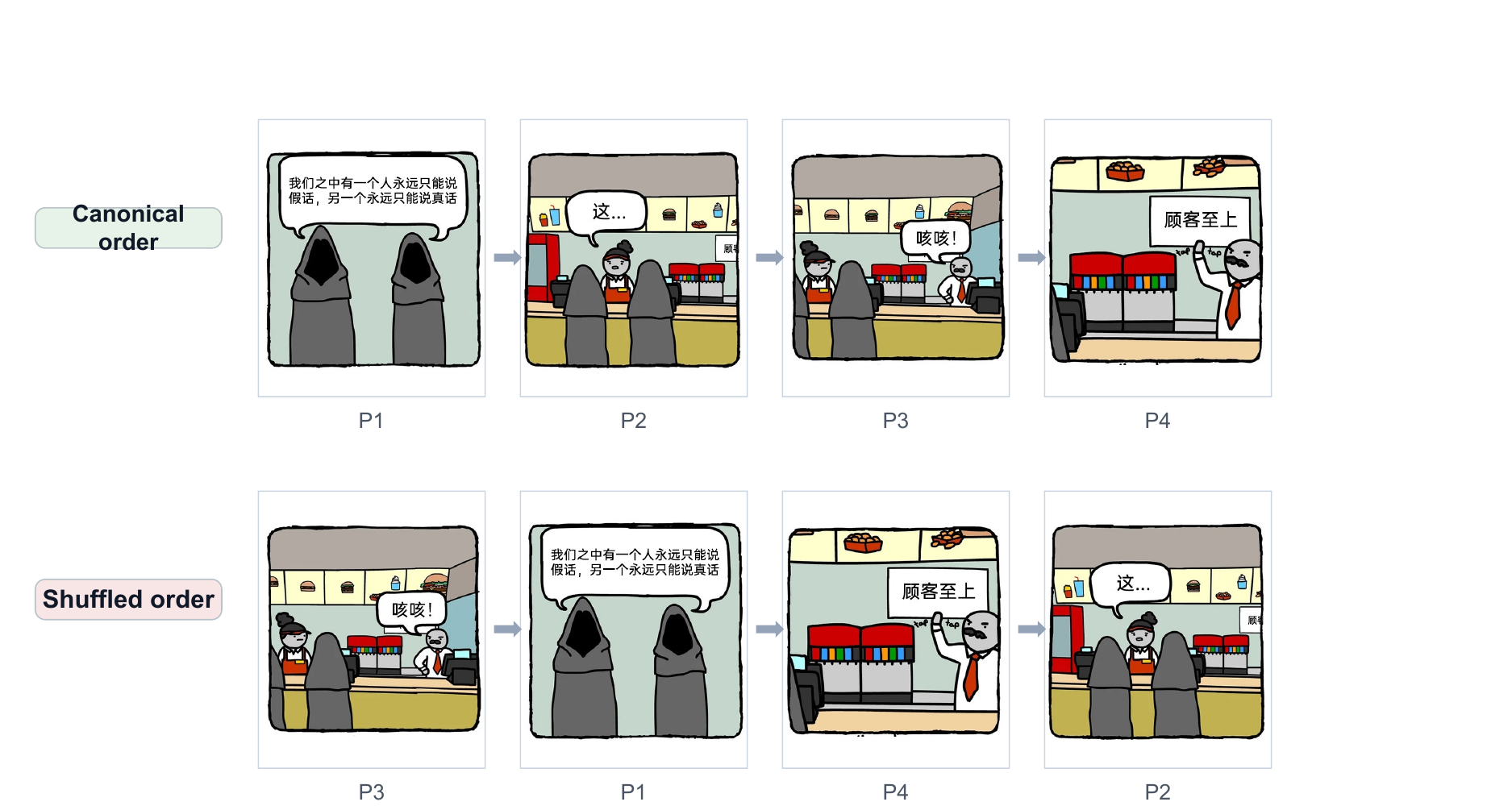}
\caption{Order matters in multi-panel memes. In the canonical order, P1 establishes a truth-versus-lie premise, P2--P3 form an interrupted conversational transition, and P4 delivers the service-style punchline ``Customer first.'' The shuffled presentation preserves the local visual content but breaks this setup-to-punchline progression, illustrating why local cue matching is insufficient for global meme understanding. The Chinese dialogue reads: P1, ``One of us can only lie forever, and the other can only tell the truth forever''; P2, ``This...''; P3, ``Oops!''; P4, ``Customer first.''}
\label{fig:teaser}
\end{figure}

We introduce \textbf{CMPM} (Chinese Multi-Panel Memes), a benchmark designed for sequence-aware vision-language evaluation.
Each sample provides cropped panels, structural type labels, ordering dependency, order-group constraints, and optional title/comment context.
Building on this resource, we define a two-layer protocol:
\textbf{Task~1} asks models to recover structure and reading order under controlled display and context conditions;
\textbf{Task~2} asks models to generate Chinese explanations that are scored by human raters on five 1--3 ordinal dimensions (visual correctness, panel coherence, humor comprehension, context handling, and faithfulness), providing a human-evaluation layer complementary to automatic LVLM metrics.

Our main contributions are:
\begin{itemize}
\item We introduce CMPM, a Chinese multi-panel meme benchmark with 1{,}214 samples and a fine-grained annotation schema that jointly represents structure, order, and context relations.
\item We design a two-layer, order-sensitive evaluation protocol spanning Task~1 panel sequencing and Task~2 human explanation ratings, including a controlled comment-aware context ablation setting (CAS).
\item Our evaluation shows that current LVLMs remain brittle under shuffled panel order, while post context yields a small average gain with both helpful and harmful instance-level effects.
\end{itemize}

\section{Related Work}
\label{sec:related}

\paragraph{Meme and humor datasets.}
Early meme corpora emphasize hate speech, offense, and sentiment in primarily English single-image settings~\cite{kiela2020hateful,suryawanshi2020multimodal,sharma2020semival}.
MemeReaCon~\cite{he2025memereacon} evaluates contextual meme understanding with post text and comments, but does not treat Chinese multi-panel \emph{sequence} structure as the primary variable.
CHIME~\cite{xie2025chime} targets phrase-based Chinese Internet memes for text-only LLM evaluation and is complementary to our visual multi-panel setting.
Table~\ref{tab:benchcmp} summarizes the positioning of CMPM against representative resources.

\paragraph{Sequential multimodal understanding.}
Order-sensitive reasoning also appears in comic understanding, visual narrative modeling, and document understanding, where meaning depends on the reading order of panels, speech bubbles, or layout regions. COMICS studies closure-driven inference across comic panels~\cite{iyyer2017comics}; CoMix includes reading order within a broader multi-task comic benchmark~\cite{vivoli2024comix}; and CHROMIC directly evaluates panel and description reordering for chronological reasoning~\cite{hou2026chromic}. ImageChain instead models image sequences as multi-turn conversations for next-scene description~\cite{villegas2026imagechain}.
These resources primarily target narrative continuation, chronological ordering, OCR-grounded reading order, or layout-level structure. CMPM isolates a different combination: order-sensitive interpretation of culturally grounded Chinese memes, where panel order interacts with punchlines, contrast, dialogue, and optional post context. CMPM is therefore complementary to sequential multimodal benchmarks rather than a duplicate of them.

\begin{table}[t]
\caption{Comparison with representative meme / multimodal resources. Context: use of post text or comments. HE: human-evaluated explanations. OS: order-sensitive panel sequencing.}
\label{tab:benchcmp}
\centering
\setlength{\tabcolsep}{2.8pt}
\begin{tabular}{lcccccc}
\toprule
Resource & Lang. & Multi-panel & OS & Context & HE & Scale \\
\midrule
Hateful Memes~\cite{kiela2020hateful} & EN & & & & & $\sim$10k \\
MemeReaCon~\cite{he2025memereacon} & EN & & & $\checkmark^*$ & & 1{,}565 \\
CHIME~\cite{xie2025chime} & ZH & & & & $\checkmark^\dagger$ & 1.5k \\
CMPM (ours) & ZH & $\checkmark$ & $\checkmark$ & $\checkmark$ & $\checkmark$ & 1{,}214 \\
\bottomrule
\end{tabular}\\[2pt]
{\footnotesize $^*$Post context for intent, not CMPM-style panel-order ablation.
$^\dagger$Text-only meme explanation (no visual panels).}
\end{table}

\paragraph{LVLM benchmarks and multimedia reasoning.}
LVLMs unify visual encoders with large language models~\cite{liu2023llava,bai2023qwen,chen2024internvl,wang2024qwen2vl}.
General suites such as MMBench~\cite{liu2024mmbench} and MMStar~\cite{chen2024mmstar}, as well as fine-grained document benchmarks such as MMDocBench~\cite{zhu2026mmdocbench}, stress perception and grounding, yet rarely isolate multi-panel \emph{meme order} as a controlled factor.
Related comic-strip and document-understanding settings likewise treat panels as spatially organized units, but generally do not isolate meme-specific reading-order recovery under controlled shuffling.
CMPM contributes a focused multimedia modeling setting: sequence-aware understanding of culturally grounded visual memes.

\paragraph{Explanation and human evaluation.}
Task~2 complements automatic LVLM metrics with human ratings of free-form Chinese explanations.

\section{Problem Definition}
\label{sec:problem}

Let a multi-panel meme be an ordered sequence $P=(p_1,\ldots,p_n)$ ($n\ge 2$) with optional post context $C$.
Gold labels include type $y_{\mathrm{type}}$, ordering dependency $y_{\mathrm{od}}\in\{\mathrm{strong},\mathrm{weak}\}$, and order groups $G$; a canonical order $\pi^\star$ is one valid linear extension of the precedence constraints encoded by $G$.
Table~\ref{tab:taxonomy} summarizes the task taxonomy.

\begin{table}[t]
\caption{CMPM task taxonomy. Primary order pool: gold-\texttt{strong} non-parallel samples with order groups.}
\label{tab:taxonomy}
\centering
\small
\begin{tabular}{L{1.05cm}L{2.25cm}L{5.35cm}L{2.25cm}}
\toprule
Task & Sub-task & Setting / $n$ & Metrics \\
\midrule
\multirow{3}{*}{Task~1}
 & Structure typing & Canonical, no comments; $N{=}1214$ & Acc., macro-/w-$F_1$ \\
 & Order recovery & Shuffled, no comments; $n{=}868$ & Group-aware order Acc. \\
 & CAS & $2{\times}2$ display$\times$context; ${\approx}252$/cell & Order Acc. \\
\midrule
Task~2
 & Explanation + human ratings & Main 64 full-context memes; paired ablation on 31 matched pairs; 3 raters & Mean Likert (5 dims, 1--3) \\
\bottomrule
\end{tabular}
\end{table}

\subsection{Task~1: Structure and Order Understanding}
\label{sec:task1}

Task~1 evaluates discriminative multimodal understanding under controlled presentations.

\paragraph{Structure typing.}
Given canonical-order panels with comments hidden, the model predicts $\hat{y}_{\mathrm{type}}$ among
\texttt{narrative}, \texttt{progressive}, \texttt{conversational}, \texttt{parallel}, and \texttt{comparison}.
We report accuracy and macro-/weighted-$F_1$.

\paragraph{Order-sensitive sequencing.}
For samples with $y_{\mathrm{od}}=\mathrm{strong}$ (and non-parallel structure with available $G$), panels are displayed under a fixed shuffle.
The model predicts a reading order $\hat{\pi}$ over the currently displayed panel indices.
The primary metric is \emph{group-aware logical-order accuracy}: $\hat{\pi}$ is correct if it is a valid linear extension of the gold precedence constraints after the canonical order is mapped to display indices; within-group permutations are allowed.

\paragraph{Context ablation setting (CAS).}
On a comment-available subset, we cross display (\texttt{correct} / \texttt{shuffled}) and context (\texttt{no\_context} / \texttt{with\_comments}).
CAS measures whether comment text helps or hurts order recovery.

\subsection{Task~2: Explanation Generation and Human Evaluation}
\label{sec:task2}

Task~2 evaluates Chinese meme explanation generation with human judgments.
Given the same meme with or without post context, a model produces a Chinese explanation $\hat{e}$.
Three raters, blinded to model identity, score each explanation separately: a rating page shows one meme and one anonymized explanation, scored on five 1--3 Likert dimensions (visual correctness, panel coherence, humor comprehension, context handling, and faithfulness).
The primary aggregate for condition comparisons is Core4, defined as the mean of Visual, Panel, Humor, and Faithfulness; Context is excluded to avoid image-only scoring bias.
On a distinct comparison page, raters holistically order five anonymous explanations; this judgment is independent of scalar scores and permits ties for similarly good or similarly poor explanations. We convert each ordering into pairwise wins and ties.
We report scalar means on 64 context-present memes (3 raters) and a paired ablation on 31 matched context-present/context-absent pairs.

\section{The CMPM Dataset}
\label{sec:dataset}

\subsection{Collection and Inclusion}

CMPM contains \textbf{1{,}214} Chinese multi-panel memes collected from major social platforms, including Weibo, Bilibili, Xiaohongshu, Tieba, and Douyin.
Each sample is stored with source metadata when available, cropped panel images ($n\ge 2$), and optional textual context.
The main set keeps only multi-panel samples with $n\ge 2$; single-image cases, including complex collages, are excluded from the main evaluation scope.
The collection pipeline filters raw posts into multi-panel candidates, crops panel images, annotates structure/order/context labels, and adjudicates gold labels used to build the Task~1 and Task~2 evaluation manifests.

\subsection{Annotation Schema}

Annotators label each meme with the following core fields:

\paragraph{Primary structure type $y_{\mathrm{type}}$.}
One of five types selected by a decision tree prioritizing global layout over local cues:
\emph{parallel} (juxtaposition/analogy),
\emph{progressive} (degree intensification),
\emph{conversational} (dialogue/chat turns),
\emph{comparison} (contrast/reversal),
and \emph{narrative} (event or causal sequence).

\paragraph{Ordering dependency $y_{\mathrm{od}}$.}
Operationalized by a shuffle test: \emph{strong} if reordering makes the meme incomprehensible or removes the punchline; \emph{weak} otherwise.

\paragraph{Order groups $G$.}
Groups encode block precedence while allowing within-group permutations (e.g., $[[1],[2,3],[4]]$).

\paragraph{Context--meme relation.}
When effective text context exists, annotators label relations such as image-dominant, text-dominant, complementary, sarcastic conflict, or unclear.

\begin{figure}[t]
\centering
\includegraphics[width=\textwidth,height=0.38\textheight,keepaspectratio]{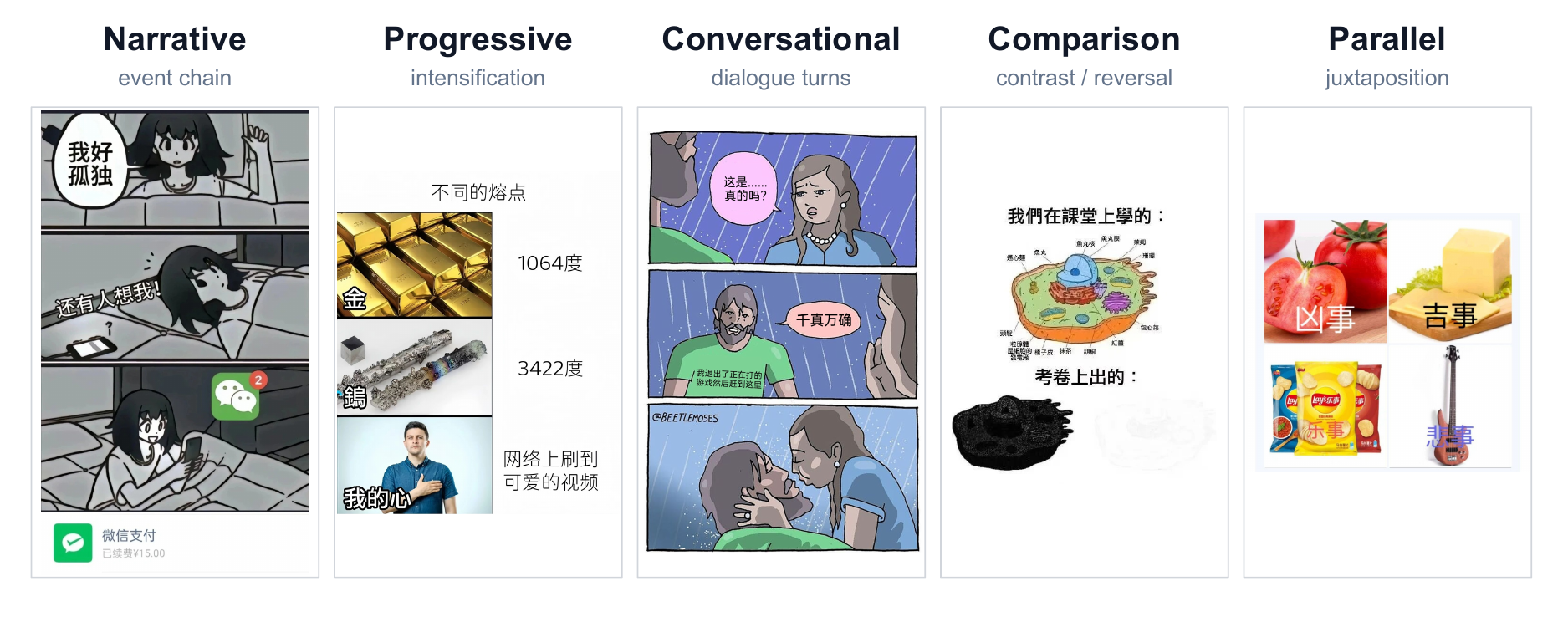}
\caption{Five primary structure types in CMPM with representative multi-panel examples (narrative, progressive, conversational, comparison, and parallel).}
\label{fig:types}
\end{figure}

\subsection{Corpus Statistics}

Table~\ref{tab:stats} reports the gold-label distribution after adjudication.
Panel counts range from 2 to 15 (411 / 257 / 332 / 129 / 47 samples for 2--6 panels; 38 samples with $\ge 7$ panels).
Here, \emph{textual context} means any non-empty title/caption/comment field attached to the post.
Among 1{,}214 memes, 322 have comment/danmaku-style context usable for CAS, of which 252 also belong to the order pool used in the context ablation.
We additionally record 991 samples with any non-empty textual field, but do not treat this broader count as equivalent to comment availability.

\begin{table}[t]
\caption{CMPM corpus statistics under final gold labels ($N=1214$).}
\label{tab:stats}
\centering
\setlength{\tabcolsep}{4pt}
\begin{tabular}{lrr|lrr}
\toprule
\textbf{Type} & $n$ & \% & \textbf{Ordering} & $n$ & \% \\
\midrule
comparison & 373 & 30.7 & strong & 868 & 71.5 \\
narrative & 319 & 26.3 & weak & 346 & 28.5 \\
conversational & 285 & 23.5 & & & \\
parallel & 137 & 11.3 & & & \\
progressive & 100 & 8.2 & & & \\
\bottomrule
\end{tabular}
\end{table}

\subsection{Annotation Quality}

Three annotators independently labeled all 1{,}214 samples under a unified guideline.
Disagreements were resolved by discussion to produce final gold labels used in all experiments.
Agreement is high for structure type (Fleiss $\kappa=0.898$) and ordering dependency (Fleiss $\kappa=0.875$).
Order-group exact match exceeds $0.97$ among non-parallel samples.
Context--meme relation is more subjective (Fleiss $\kappa=0.499$), which we treat as an auxiliary label.

\section{Evaluation Protocol}
\label{sec:protocol}

\subsection{Models}

We evaluate five LVLMs under a shared manifest so that display order, context strings, and prompts are model-agnostic:
\begin{itemize}
\item \textbf{Open-source:} InternVL3.5-8B~\cite{opengvlab2026internvl35}, Qwen3.5-9B~\cite{qwen2026qwen35}, and GLM-4.1V-9B-Thinking~\cite{zai2026glm41v};
\item \textbf{Closed-source:} GPT-5.5~\cite{openai2026gpt55} and Gemini~3.1~Pro~\cite{google2026gemini31}.
\end{itemize}
We do not fine-tune any model on CMPM.
Closed-source models are evaluated under the same final shared manifest as open models.

\subsection{Task~1 Protocol Details}

\paragraph{Type prediction.}
One request per meme: panels in canonical order, no comments ($N=1214$).

\paragraph{Order prediction.}
The order pool contains gold-\texttt{strong} non-parallel samples with valid order groups ($N=868$).
The \textbf{primary} leaderboard condition is shuffled display + no comments (shared seed $=42$).
Hard cases with $\ge 7$ panels are retained and tagged separately.
The generation manifest has 1{,}624 unique requests per model after merging primary and CAS conditions.

\paragraph{CAS.}
On 252 comment-available order-pool memes, we evaluate the $2\times 2$ grid of display $\times$ context and report group-aware logical-order accuracy for each condition in Table~\ref{tab:cas}.

\paragraph{Metrics.}
For ordering we report group-aware logical-order accuracy on parseable outputs.
For typing we report accuracy, macro-$F_1$, and weighted-$F_1$, plus parse rate.

\subsection{Task~2 Protocol Details}

Models generate explanations with or without available post text. Scalar scoring uses randomized blind pages, each showing one meme and one anonymized candidate; a separate randomized comparison page shows five candidates for holistic ranking. Three independent non-author annotators received common rubric training and independently completed both judgments with model identities and other ratings hidden. Holistic ties are allowed for similarly good or similarly poor explanations and are expanded into ten pairwise wins and ties.
The main aggregate contains 64 unique context-present memes selected from a stratified 400-meme generation pool (12--13 per structure type), balancing type, panel count, context availability, and annotation cost.
The 31-pair ablation was drawn from the same annotated candidate pool as the main set, with each meme evaluated under both context conditions. Its primary metric is Core4 (Table~\ref{tab:task2_ablation}).
Together, the evaluation covers 67 unique IDs and 98 meme-condition pages: 490 explanations, 1{,}470 explanation-level judgments, 7{,}350 scalar cells, and 294 holistic rankings converted into 2{,}940 pairwise outcomes.
This dense design targets controlled within-meme model comparison rather than population-level estimation. Main-set score agreement is moderate (Krippendorff ordinal $\alpha\approx 0.48$).

\section{Experiments}
\label{sec:exp}

Unless noted, all five models are scored against final gold labels under the shared Task~1 manifests.
Task~2 human ratings use the completed three-annotator blind evaluation protocol (Sect.~\ref{sec:protocol}); holistic rankings are reported as pairwise outcomes.

\subsection{Task~1: Primary Order Recovery}

Table~\ref{tab:order_primary} reports group-aware logical-order accuracy on the primary condition (strong, shuffled, no context).
The two italicized rows are weak references evaluated with the same group-aware parser and metric: \texttt{Display-order} follows the presented panel order, while \texttt{Random} samples one uniformly random permutation per item (seed $=42$).
\textbf{Key findings:}
Open-source models remain weak on shuffled order recovery: Qwen3.5-9B is best at 27.4\%, while GLM-4.1V-9B-Thinking and InternVL3.5-8B reach 15.6\% and 6.1\%, respectively.
Closed-source models are substantially stronger---Gemini~3.1~Pro reaches 75.2\% and GPT-5.5 57.8\%---but both still leave considerable room for improvement.
Across both model families, canonical-display accuracy does not transfer reliably to shuffled inputs (Table~\ref{tab:cas}): the correct-versus-shuffled gap persists, and Gemini's shuffled cells (73.0\% / 71.3\%) remain clearly below its correct-display cells (90.5\% / 94.8\%), confirming persistent order sensitivity.

\begin{table}[!ht]
\caption{Task~1 primary order recovery (gold-\texttt{strong}, shuffled, no context). Acc.\ denotes group-aware logical-order accuracy (\%).}
\label{tab:order_primary}
\centering
\begin{tabular}{lcc}
\toprule
Model & $n$ scored & Acc.\ (\%) \\
\midrule
InternVL3.5-8B & 863 & 6.1 \\
Qwen3.5-9B & 865 & 27.4 \\
GLM-4.1V-9B-Thinking & 865 & 15.6 \\
GPT-5.5 & 868 & 57.8 \\
Gemini~3.1~Pro & 868 & 75.2 \\
\midrule
\textit{Display-order} & 868 & 0.0 \\
\textit{Random (seed=42)} & 868 & 19.4 \\
\bottomrule
\end{tabular}\\[2pt]
{\footnotesize Model rows use parseable outputs; the smaller open-model counts reflect malformed or incomplete order outputs, including failures on longer inputs (especially beyond four panels).}
\end{table}

\subsection{Task~1: Structure Typing}

Table~\ref{tab:type} summarizes type prediction on all 1{,}214 memes.
Open-source models remain in the mid-50\% range, with the best accuracy at 56.1\% and macro-$F_1$ scores of 41.8--50.7. Closed-source models perform better: GPT-5.5 reaches 76.3\% accuracy / 74.9 macro-$F_1$, while Gemini~3.1~Pro reaches 69.7\% / 67.9. With parse rates above 99.7\%, the bottleneck is semantic structure recognition rather than output validity, especially at the narrative--conversational and parallel--comparison boundaries (Sect.~\ref{sec:analysis}).

\begin{table}[!ht]
\caption{Task~1 structure typing ($N=1214$, canonical order, no context). Metrics in \%.}
\label{tab:type}
\centering
\setlength{\tabcolsep}{3.5pt}
\begin{tabular}{lcccc}
\toprule
Model & Acc. & Macro-$F_1$ & W.-$F_1$ & Parse \\
\midrule
InternVL3.5-8B & 53.6 & 41.8 & 48.7 & 99.7 \\
Qwen3.5-9B & 56.1 & 47.4 & 51.2 & 100.0 \\
GLM-4.1V-9B-Thinking & 55.8 & 50.7 & 52.5 & 99.8 \\
GPT-5.5 & 76.3 & 74.9 & 76.4 & 100.0 \\
Gemini~3.1~Pro & 69.7 & 67.9 & 67.0 & 100.0 \\
\bottomrule
\end{tabular}\\[2pt]
\end{table}

\subsection{Context Ablation (CAS)}

Table~\ref{tab:cas} reports logical-order accuracy on the CAS $2\times 2$ grid.
Shuffled displays remain substantially harder than correct displays, and comments do not close this gap for either open- or closed-source models. Closed-source models are more robust under shuffling, with GPT-5.5 at approximately 51\% and Gemini~3.1~Pro at 71--73\%, but both remain order-sensitive. Comment effects are inconsistent: InternVL improves from 4.8\% to 6.8\% under shuffle, Qwen stays near 23\%, and Gemini declines slightly with comments. Thus, post-context offers limited, model-specific assistance but cannot substitute for visual sequence reasoning.

\begin{table}[!ht]
\caption{CAS logical-order accuracy (\%). Each cell uses comment-available order-pool memes ($n\approx 250$--$252$ scored).}
\label{tab:cas}
\centering
\setlength{\tabcolsep}{2.5pt}
\begin{tabular}{lcccc}
\toprule
& \multicolumn{2}{c}{Correct display} & \multicolumn{2}{c}{Shuffled display} \\
\cmidrule(lr){2-3}\cmidrule(lr){4-5}
Model & no ctx & +cmts & no ctx & +cmts \\
\midrule
InternVL3.5-8B & 92.8 & 85.6 & 4.8 & 6.8 \\
Qwen3.5-9B & 85.7 & 88.9 & 23.5 & 23.1 \\
GLM-4.1V-9B-Thinking & 97.2 & 96.0 & 10.4 & 13.5 \\
GPT-5.5 & 92.9 & 91.7 & 51.2 & 51.2 \\
Gemini~3.1~Pro & 90.5 & 94.8 & 73.0 & 71.3 \\
\bottomrule
\end{tabular}
\end{table}

\subsection{Difficulty by Length}

On the primary condition, long strips are markedly harder than shorter ones: GPT-5.5 drops from 59.1\% on \texttt{main\_le6} to 16.0\% on \texttt{hard\_7plus}, and Gemini~3.1~Pro from 76.7\% to 24.0\%; open models likewise remain near floor on hard items (InternVL 4.8\%, Qwen 8.0\%, GLM 8.3\%).
Across all scored Task~1 order requests, the \texttt{panel\_4plus} bucket is also relatively difficult (GPT-5.5 63.7\%, Gemini 74.2\%), trailing the corresponding 2-/3-panel overall rates.

\subsection{Task~2: Explanation Generation}

Table~\ref{tab:task2} reports five-dimension mean scores on the main set (64 unique context-present memes).
As a secondary consistency check, each independently collected five-way ordering is expanded into pairwise wins and ties rather than reconstructed from summed Likert scores. Table~\ref{tab:task2_ablation} reports the paired Core4 ablation separately. These pairwise outcomes recover the broad tiers in the scalar ratings: Gemini~3.1~Pro and GPT-5.5 are preferred over the open models, while GLM is slightly favored over Qwen in their direct comparison. The gap is substantial (Gemini $\approx$2.7--2.8 vs.\ InternVL $\approx$1.9--2.2). Adding post context raises average Core4 only from 2.30 to 2.36 ($\Delta{=}+0.06$); using a per-meme change of at least 0.1, context helps 13 pairs, hurts 10, and has a small or mixed effect on 8. Context is therefore conditionally useful rather than universally beneficial, and inter-annotator agreement is moderate (ordinal $\alpha\approx0.48$).

\begin{table}[!ht]
\caption{Task~2 main scalar evaluation (64 unique context-present memes). Mean of five Likert dimensions (1--3) per annotator. Higher is better; relative preferences are derived separately from holistic rankings.}
\label{tab:task2}
\centering
\setlength{\tabcolsep}{3.5pt}
\begin{tabular}{lccc}
\toprule
Model & ann\_A & ann\_B & ann\_C \\
\midrule
InternVL3.5-8B & 1.91 & 2.03 & 2.16 \\
Qwen3.5-9B & 2.00 & 2.08 & 2.33 \\
GLM-4.1V-9B-Thinking & 2.03 & 2.36 & 2.47 \\
GPT-5.5 & 2.43 & 2.49 & 2.65 \\
Gemini~3.1~Pro & 2.78 & 2.71 & 2.79 \\
\bottomrule
\end{tabular}\\[2pt]
\end{table}

\begin{table}[!ht]
\caption{Task~2 context ablation (ABC combined; 31 matched pairs from the curated Task~2 evaluation pool; $n{=}93$ ratings per model--condition cell and $n{=}465$ for the overall aggregate). Core4 $=$ mean of Visual, Panel, Humor, and Faithfulness (excludes Context).}
\label{tab:task2_ablation}
\centering
\setlength{\tabcolsep}{3.5pt}
\begin{tabular}{lccc}
\toprule
Model & \texttt{full\_context} & \texttt{image\_only} & $\Delta$ \\
\midrule
InternVL3.5-8B & 2.03 & 1.87 & +0.16 \\
Qwen3.5-9B & 2.09 & 2.09 & +0.00 \\
GLM-4.1V-9B-Thinking & 2.38 & 2.37 & +0.01 \\
GPT-5.5 & 2.52 & 2.49 & +0.03 \\
Gemini~3.1~Pro & 2.77 & 2.69 & +0.08 \\
\midrule
Overall & 2.36 & 2.30 & +0.06 \\
\bottomrule
\end{tabular}\\[2pt]
\end{table}

\section{Analysis and Discussion}
\label{sec:analysis}

\subsection{Order Sensitivity}

The dominant failure mode is \emph{order blindness under shuffle}.
Open models often emit plausible left-to-right sequences that ignore causal, contrastive, or dialogue constraints after permutation. Closed-source models narrow but do not eliminate the gap (Table~\ref{tab:cas}): GPT-5.5 remains near 51\% under shuffle, while Gemini~3.1~Pro remains well below its correct-display scores. Thus canonical accuracy can reflect exploitation of the presented order, whereas shuffled evaluation tests whether panel recognition is integrated into reliable sequence reasoning.

\subsection{Structure Confusions}

Errors concentrate at semantic boundaries, especially narrative--conversational and parallel--comparison; the rarer progressive class is often absorbed into them. This mirrors annotation boundary cases and shows that LVLMs do not consistently encode structural distinctions among layouts.

\subsection{Does Comment Context Help?}

Comments are not a universal remedy: they help when restating a punchline but can mislead when ironic, referential, or unrelated to panel order. Visual sequence reasoning remains necessary.

\subsection{Qualitative Error Modes}

\begin{figure}[!h]
\centering
\includegraphics[width=\textwidth,trim=0pt 22pt 0pt 22pt,clip]{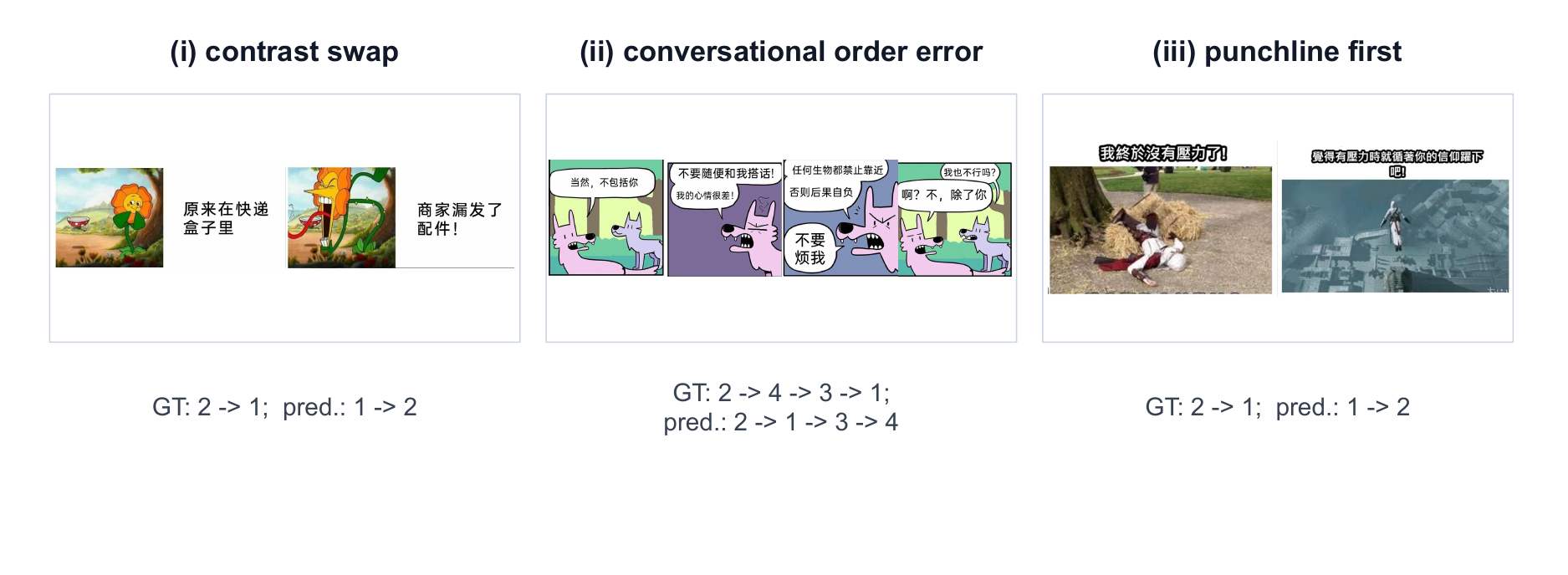}
\vspace{-0.5em}
\caption{Illustrative Task~1 order-recovery errors under shuffled display. Numbers are shuffled display indices; each subfigure contrasts gold and observed orders.}
\label{fig:errors}
\end{figure}

Figure~\ref{fig:errors} highlights three recurring failure modes: contrast
swaps reverse the intended progression, conversational errors misorder reply
turns, and punchline-first errors promote a salient final panel. Together,
these cases show that order recovery requires discourse relations, not panel
recognition alone.

\subsection{Limitations}

CMPM focuses on Chinese multi-panel social memes and may not cover other
cultures or single-image templates. Existing LVLM benchmarks more often use
single images or document pages, so multi-panel sequence settings remain less
common. The study evaluates frozen checkpoints and proprietary APIs, making
results sensitive to prompts, decoding settings, and future model updates.
Task~2 agreement is moderate ($\alpha\approx 0.48$), and its curated subset
targets controlled comparison rather than population-level estimation.
Release will remove identifying metadata and honor takedown requests.

\section{Conclusion}
\label{sec:conclusion}

We introduced CMPM, a 1,214-sample Chinese multi-panel meme benchmark for
order-sensitive reasoning. Shuffling reveals order blindness: Gemini~3.1~Pro
reaches 75.2\% on the primary condition, while open models remain lower.
Structure typing is more reliable than boundary-sensitive ordering; comment
gains are heterogeneous. Task~2 favors Gemini~3.1~Pro and GPT-5.5, with
moderate agreement cautioning against a single score. Future work has three
levels: expand data across languages, cultures, panel counts, and adversarial
permutations; add panel-aware representations, cross-panel memory, and
lightweight order-consistency adapters; and develop equivalence-aware,
attention-grounded evaluation across model versions.

\begingroup
\sloppy
\bibliographystyle{splncs04}
\bibliography{refs}
\endgroup
\end{document}